%% file: arxiv17_tgifqa.tex
\documentclass[10pt,twocolumn,letterpaper]{article}

\include{mymacros}

\usepackage{cvpr}
\usepackage{times}
\usepackage{epsfig}
\usepackage{graphicx}
\usepackage{amsmath}
\usepackage{amssymb}
\usepackage{paralist}

\usepackage[dvipsnames]{xcolor}

\usepackage[breaklinks=true,bookmarks=false,colorlinks]{hyperref}

\cvprfinalcopy %

\def\cvprPaperID{1002} %

\ifcvprfinal\fi

\begin{document}

\title{TGIF-QA: Toward Spatio-Temporal Reasoning in Visual Question Answering}

\author{
    Yunseok Jang$^1$, Yale Song$^2$, Youngjae Yu$^1$,
    Youngjin Kim$^1$, Gunhee Kim$^1$ \\
    $^1$Seoul National University, $^2$Yahoo Research\\
    {\tt\footnotesize $^1$\{yunseok.jang, gunhee\}@snu.ac.kr, \{yj.yu, youngjin.kim\}@vision.snu.ac.kr}
    \quad \tt\footnotesize $^2$yalesong@yahoo-inc.com\\
    \tt\normalsize \href{http://vision.snu.ac.kr/projects/tgif-qa}{\color{magenta}http://vision.snu.ac.kr/projects/tgif-qa}
}

\maketitle

\begin{abstract}
Vision and language understanding has emerged as a subject undergoing intense study in Artificial Intelligence. Among many tasks in this line of research, visual question answering (VQA) has been one of the most successful ones, where the goal is to learn a model that understands visual content at region-level details and finds their associations with pairs of questions and answers in the natural language form. Despite the rapid progress in the past few years, most existing work in VQA have focused primarily on images. In this paper, we focus on extending VQA to the video domain and contribute to the literature in three important ways. First, we propose three new tasks designed specifically for video VQA, which require spatio-temporal reasoning from videos to answer questions correctly. Next, we introduce a new large-scale dataset for video VQA named TGIF-QA that extends existing VQA work with our new tasks. Finally, we propose a  dual-LSTM based approach with both spatial and temporal attention, and show its effectiveness over conventional VQA techniques through empirical evaluations. 
\end{abstract} 

\begin{figure}[ht]
    \centering
    \includegraphics[width=\linewidth]{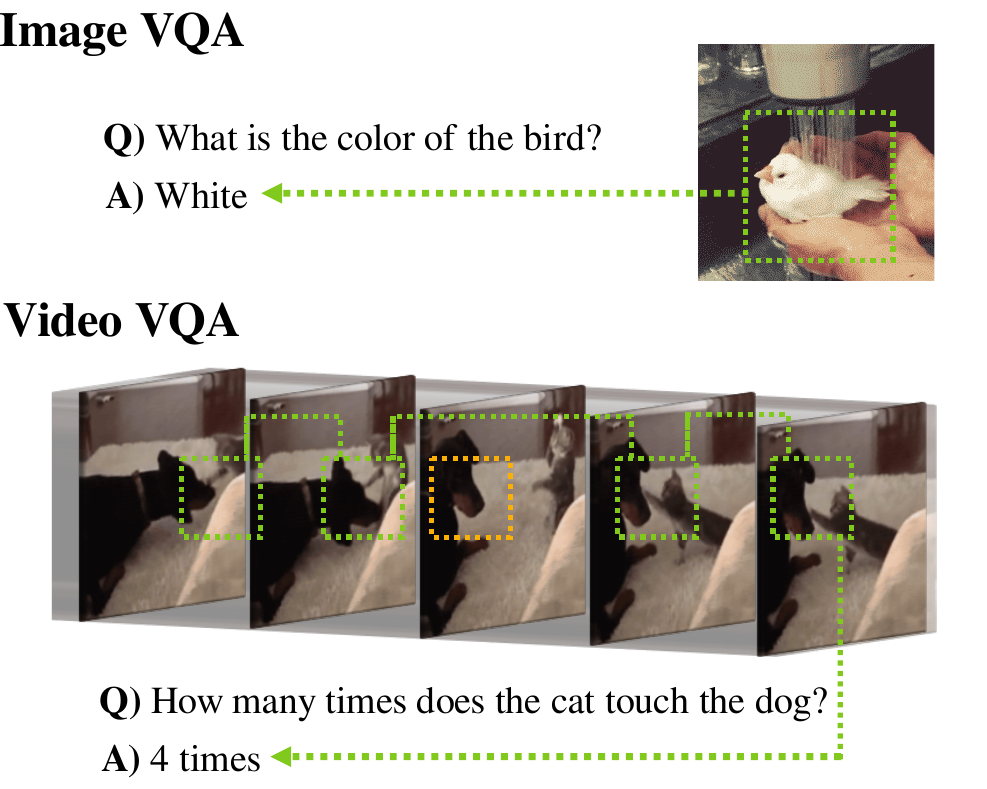}
    \caption{Much of conventional VQA tasks focus on reasoning from images (top). This work proposes a new dataset with tasks designed specifically for video VQA that requires spatio-temporal reasoning from videos to answer questions correctly (bottom).}
    \label{fig:difference_image_and_video}
    \vspace{-9pt}
\end{figure}

\begin{figure*}[ht]
    \centering
    \includegraphics[width=0.98\linewidth]{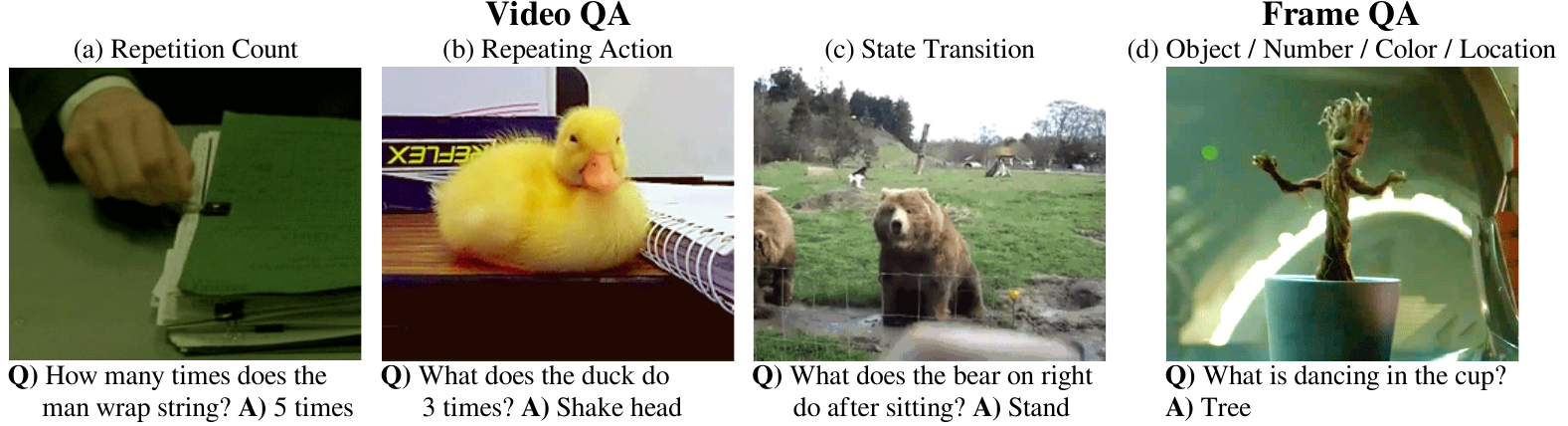}
    \caption{Our TGIF-QA dataset introduces three new tasks for video QA, which require spatio-temporal reasoning from videos (\eg (a) repetition count, (b) repeating action, and (c) state transition). 
It also includes frame QA tasks that can be answered from one of frames.}
    \label{fig:example_questions}
    \vspace{-6pt}
\end{figure*}

\section{Introduction}
\label{sec:introduction}
Vision and language understanding has emerged as a subject undergoing intense study in Artificial Intelligence. Among many tasks in this line of research, visual question answering (VQA) has been one of the most successful ones, where the goal is to learn a model that understands visual content at region-level details and finds their associations with pairs of questions and answers in the natural language form~\cite{antol-iccv-2015}. Part of the reasons for the success of VQA is that there exists a number of large-scale datasets with well-defined tasks and evaluation protocols~\cite{antol-iccv-2015,malinowski-nips-2014,ren-nips-2015,zhu-cvpr-2016}, which provided a common ground to researchers to compare their methods in a controlled setting.

While we have seen a rapid progress in video analysis~\cite{karpathy-cvpr-2014,srivastava-icml-2015,tran-iccv-2015}, most existing work in VQA have focused primarily on images. We believe that the limited progress in video VQA, compared to its image counterpart, is due in part to the lack of large-scale datasets with well-defined tasks. Some early attempts have been made to fill this gap by introducing datasets that leverage movie data~\cite{rohrbach-arxiv-2016,tapaswi-cvpr-2016}, focusing on storyline comprehension either from short video clips~\cite{rohrbach-arxiv-2016} or from movies and scripts~\cite{tapaswi-cvpr-2016}. However, existing question-answer pairs are either an extension to the conventional image VQA tasks, \eg, by adding action verbs as the new answer type~\cite{rohrbach-arxiv-2016} to the existing categories of object, number, color, and location~\cite{ren-nips-2015}, or require comprehensive understanding of long textual data, \eg, movie scripts~\cite{tapaswi-cvpr-2016}.

In this paper, we contribute to the literature in VQA in three important ways. First, we propose three new tasks designed specifically for video VQA, which require spatio-temporal reasoning from videos to answer questions correctly. Next, we introduce a new large-scale dataset for video VQA that extends existing work in image VQA with our new tasks. Finally, we propose a dual-LSTM based approach with an attention mechanism to solve our problem, and show its effectiveness over conventional VQA techniques through empirical evaluations. Our intention is not to compete with existing literature in VQA, but rather to complement them by providing new perspectives on the importance of spatio-temporal reasoning in VQA. %

Our design of video VQA tasks is inspired by existing works in video understanding, e.g., repetition counting~\cite{levy-iccv-2015} and state transitions~\cite{isola-cvpr-2015}, intending to serve as a bridge between video understanding and video VQA. We define three tasks: (1) count the number of repetitions of a given action; (2) detect a repeating action given its count; and (3) identify state transitions, \ie, what has happened before or after a certain action state. As illustrated in Figure~\ref{fig:example_questions}, solving our tasks requires comprehensive spatio-temporal reasoning from videos, an ideal scenario for evaluating video analysis techniques. In addition to our new tasks, we also include the standard image VQA type tasks by automatically generating question-answer pairs from video captions~\cite{ren-nips-2015}. Following the existing work in VQA, we formulate our questions as either open-ended or multiple choice. This allows us to take advantage of well-defined evaluation protocols.

To create a benchmark for our tasks, we collected a new dataset for video VQA based on the Tumblr GIF (TGIF) dataset~\cite{li-cvpr-2016}, which was originally proposed for video captioning. The TGIF dataset utilizes animated GIFs as their visual data, which have recently emerged as an attractive source of data in computer vision~\cite{gygli-cvpr-2016,li-cvpr-2016} due to their concise format and cohesive storytelling nature~\cite{bakhshi-chi-2016}; this makes it especially ideal for vision and language understanding. We therefore extend the TGIF dataset to the VQA domain, adding 165K QA pairs from 72K animated GIFs from the TGIF dataset; we name our dataset \textit{TGIF-QA}. %

The current state-of-the-art in VQA have focused on finding visual-textual associations from images~\cite{antol-iccv-2015,ren-nips-2015}, employing a spatial attention mechanism to learn ``where to look'' in an image given the question~\cite{fukui-emnlp-2016,kim-nips-2016}. While existing techniques demonstrated impressive performance on image VQA, they are inadequate for the video domain because a video contains visual information both in spatial and temporal dimensions, requiring an appropriate spatio-temporal reasoning mechanism. In this work, we leverage spatio-temporal information from video by employing LSTMs not only for the QA pairs, as in the previous works, but also for the video input. We also evaluate spatial and temporal attention mechanisms to selectively attend to specific parts of a video. We discuss various design considerations and report empirical results in Section~\ref{sec:experiments}. 

In this updated version of the paper, we extend our dataset by collecting more question and answer pairs (the total count has increased from 104K to 165K) and update all relevant statistics, including Table \ref{tab:dataset}. Also, we retake all the evaluations with the extended dataset and include language-only baseline results in Table \ref{tab:experiment_results}.

To summarize, our major contributions include:
\begin{enumerate}
\item We propose three new tasks designed specifically for video VQA, which require spatio-temporal reasoning from videos to answer questions correctly.
\item We introduce a new dataset, TGIF-QA, that consists of 165K QA pairs from 72K animated GIFs.
\item We propose a dual-LSTM based approach with an attention mechanism to solve our video QA tasks.
\item Code and the dataset are available on our project page.
\end{enumerate}

\section{Related Works}
\label{sec:related_works}
VQA is a relatively new problem domain first introduced by Malinowski~\etal~\cite{malinowski-nips-2014} and became popularized by Antol~\etal~\cite{antol-iccv-2015}. Despite its short history, there has been a flourishing amount of research produced within the past few years~\cite{ferraro-emnlp-2015,bernardi-jair-2016}. Here, we position our research and highlight key differences compared to previous work in VQA.

\textbf{Datasets.} Most existing VQA datasets are image-centric, \eg, DAQUAR~\cite{malinowski-nips-2014}, abstract scenes~\cite{lin-cvpr-2015}, VQA~\cite{antol-iccv-2015}, Visual Madlibs~\cite{yu-iccv-2015}, DAQUAR-Consensus~\cite{malinowski-iccv-2015}, FM-IQA~\cite{gao-nips-2015}, COCO-QA~\cite{ren-nips-2015}, and Visual7W~\cite{zhu-cvpr-2016}. Also, appearing in the same proceedings are CLEVR~\cite{johnson-cvpr-2017}, VQA2.0~\cite{goyal-cvpr-2017}, and Visual Dialog~\cite{das-cvpr-2017}, which all address image-based VQA. Our work extends  existing works to the video domain, creating QA pairs from short video clips rather than static images.

There have been some recent efforts to create video VQA datasets based on movies. Rohrbach~\etal~\cite{rohrbach-arxiv-2016} extended the LSMDC movie description dataset~\cite{rohrbach-arxiv-2016} to the VQA domain. Similarly, Tapaswi~\etal~\cite{tapaswi-cvpr-2016} introduced the MovieQA dataset by leveraging movies and movie scripts. Our work contributes to this line of research, but instead of restricting the source of video to the movie clips, here we leverage animated GIFs from the Internet, which have concise format and deliver cohesive visual stories~\cite{bakhshi-chi-2016,li-cvpr-2016}.

\textbf{Tasks.} Existing QA pairs in the VQA literature have one of the following forms: open-ended and multiple choice; we consider fill-in-the-blank as a special case of the open-ended form. Open-ended questions provide either a complete or incomplete sentence and the system must guess the correct answer word. Multiple choice questions, on the other hand, provide a number of answer candidates, either as texts~\cite{antol-iccv-2015} or bounding boxes~\cite{zhu-cvpr-2016}, and the system must choose the correct one. Our dataset contains questions in the open-ended and multiple choice forms.%

Most existing VQA tasks are image-centric and thus ask questions about visual concepts that appear only in images, \eg, objects, colors, and locations~\cite{antol-iccv-2015}. In the video domain, the LSMDC-QA dataset~\cite{rohrbach-arxiv-2016} introduced the movie fill-in-the-blank task by adding action verbs to the answer set, requiring spatio-temporal reasoning from videos at the sequence level (similar to action recognition). Our tasks also require spatio-temporal reasoning from videos, but at the frame level -- counting the number of repetitions and memorizing state transitions from a video requires more comprehensive spatio-temporal reasoning.

The MovieQA dataset~\cite{tapaswi-cvpr-2016} introduced an automatic story comprehension task from video and movie script. The questions are designed to require comprehensive visual-textual understanding of a movie synopsis, to the level of details of proper nouns (\eg, names of characters and places in a movie). Compared to the MovieQA dataset, our task is on spatio-temporal reasoning rather than story comprehension, and we put more focus on understanding visual signals (animated GIFs) rather than textual signals (movie scripts).

\textbf{Techniques.} Most existing techniques in VQA are designed to solve image VQA tasks. Various techniques have demonstrated promising results, such as the compositional model~\cite{andreas-cvpr-2016} and the knowledge-based model~\cite{wu-cvpr-2016}. The current state-of-the-art techniques employ a spatial attention mechanism with visual-textual joint embedding~\cite{fukui-emnlp-2016,kim-nips-2016}. Our work extends this line of work to the video domain, by employing spatial and temporal attention mechanisms to solve video VQA tasks. 

There are very few approaches designed specifically to solve video VQA. Yu~\etal~\cite{yu-arxiv-2016} used LSTMs to represent both videos and QA pairs and adopted a semantic attention mechanism~\cite{you-cvpr-2016} on both input word representation and output word prediction. We also use LSTMs to represent both videos and QA pairs, with a different attention mechanism to capture complex spatio-temporal patterns in videos. To the best of our knowledge, our model is the first to leverage temporal attention for video VQA tasks, which turns out to improve the QA performance in our experiments. 

\begin{table}[t]
\centering
\footnotesize
\setlength{\tabcolsep}{2.5pt}
\begin{tabular}{|l|c|ccc|ccc|}
\hline
\multicolumn{2}{|c|}{\multirow{2}*{Task}}           & \multicolumn{3}{|c|}{\# QA pairs} & \multicolumn{3}{|c|}{\# GIFs} \\\cline{3-8}
\multicolumn{2}{|c|}{}                              & Train     & Test      & Total     & Train     & Test      & Total \\\hline
\multirow{3}{0.4cm}{Video QA}   & Rep. Count        & 26,843    & 3,554     & 30,397    & 26,843    & 3,554     & 30,397 \\
                                & Rep. Action       & 20,475    & 2,274     & 22,749    & 20,475    & 2,274     & 22,749 \\
                                & Transition        & 52,704    & 6,232     & 58,936    & 26,352    & 3,116     & 29,468 \\\hline
                                & Object            & 16,755    & 5,586     & 22,341    & 15,584    & 3,209     & 18,793 \\
Frame                           & Number            & 8,096     & 3,148     & 11,244    & 8,033     & 1,903     & 9,936  \\
QA                              & Color             & 11,939    & 3,904     & 15,843    & 10,872    & 3,190     & 14,062 \\
                                & Location          & 2,602     & 1,053     & 3,655     & 2,600     & 917       & 3,517 \\\hline
\multicolumn{2}{|c|}{Total}                         & 139,414   & 25,751    & 165,165   & 62,846    & 9,575     & 71,741     \\\hline
\end{tabular}
\vspace{3pt}
\caption{Statistics of our dataset, organized into different tasks.}
\label{tab:dataset}
\vspace{-6pt}
\end{table}

\section{TGIF-QA Dataset}
\label{sec:dataset}

Our dataset consists of 165,165 QA pairs collected from 71,741 animated GIFs. We explain our new tasks designed for video VQA and present the data collection process. %

\subsection{Task Definition} 
\label{sec:types}

We introduce four task types used in our dataset. Three of them are new and unique to the video domain, including:

\textbf{Repetition count.} One task that is truly unique to videos would be counting the number of repetitions of an action. We define this task as an open-ended question about counting the number of repetitions of an action, \eg, Figure~\ref{fig:example_questions} (a). There are 11 possible answers (from 0 to 10+).

\textbf{Repeating action.} A companion to the above, this task is defined as a multiple choice question about identifying an action that has been repeated in a video, \eg, Figure~\ref{fig:example_questions} (b). We provide 5 options to choose from.

\textbf{State transition.} Another task unique to videos is asking about transitions of certain states, including facial expressions (\eg, from happy to sad), actions (\eg, from running to standing), places (\eg, from the table to the floor), and object properties (\eg, from empty to full). We define this task as a multiple choice question about identifying the state before (or after) another state, \eg, Figure~\ref{fig:example_questions} (c). We provide 5 options to choose from.

The three tasks above require analyzing multiple frames of a video; we refer to them collectively by \textbf{video QA}. 

Besides our three video QA tasks, we also include another one, which we call \textbf{frame QA} to highlight the fact that questions in this task can be answered from one of the frames in a video. Depending on the video content, it can be any frame or one particular from of a video. For this task, we leverage the video captions provided in the TGIF dataset~\cite{li-cvpr-2016} and use the NLP-based technique proposed in Ren~\etal~\cite{ren-nips-2015} to generate QA pairs automatically from the captions. This task is defined as an open-ended question about identifying the best answer (from a dictionary of words of type object, number, color, and location) given a question in a complete sentence, \eg, Figure~\ref{fig:example_questions} (d).

\begin{table}[t]
\small
\centering
\begin{tabular}{|c|l|c|}
\hline
 Task & Question & Answer \\ \hline
Repetition  & How many times does the                       & \multirow{2}*{[\texttt{\#Repeat}]}    \\
count       & [\texttt{SUB}] [\texttt{VERB}] [\texttt{OBJ}] ? &                                         \\\hline 
Repeating   & What does the [\texttt{SUB}] do               &   \multirow{2}*{[\texttt{VERB}] [\texttt{OBJ}]}                       \\
action      & [\texttt{\#Repeat}] times ?                   &                           \\\hline
            & What does the [\texttt{SUB}] do               &   [\texttt{Previous}                  \\
State       & before [\texttt{Next state}] ?                &   \texttt{state}]                     \\\cline{2-3}
transition  & What does the  [\texttt{SUB}] do              &   [\texttt{Next}                      \\
            & after [\texttt{Previous state}] ?             &   \texttt{state}]                     \\\hline
\end{tabular}
\vspace{6pt}
\caption{Templates used for creating video QA pairs.}  
\label{tab:question_template} 
\vspace{-6pt}
\end{table}

\subsection{QA Collection} 
\label{sec:collection}

For the frame QA, we use the same setup of Ren~\etal~\cite{ren-nips-2015} and apply their method on the captions provided in the TGIF dataset~\cite{li-cvpr-2016}. As shown in Table~\ref{tab:dataset}, this produced a total of 53,083 QA pairs from 39,479 GIFs.  %
For the video QA, we generate QA pairs by using a combination of crowdsourcing and template-based approach. This produced a total of 112,082 QA pairs from 53,247 GIFs. %

\textbf{Crowdsourcing.} We conducted two crowdsourcing studies, collecting the following information:
\begin{itemize}%
\item Repetition: subject, verb, object, and the number of repetitions (from 2 to 10+ times) for a repeating action.
\item State transition: subject, transition type (one of facial expression, action, place, or object property), previous state, next state for the changed states, if any.
\end{itemize}%
We used drop-down menus to collect answers for the number of repetitions and the transition type, and used text boxes for all the others. A total of 595 workers have participated and were compensated by 5 cents per video clip. 

\textbf{Quality control.} Our task includes many free-form input; proper quality control is crucial. Inspired by Li~\etal~\cite{li-cvpr-2016}, we filter out suspiciously negligent workers by automatic validation. Specifically, we collect a small set of video clips (159 for repetition and 172 for state transition) as the validation set, and manually annotate each example with a set of appropriate answers; we consider those the gold standard. We then include one of the validation samples to each main task and check if a worker answers it correctly by matching their answers to our gold standard set. We reject the answers from workers who fail to pass our validation, and add those workers to our blacklist so that they cannot participate in other tasks. We regularly reviewed rejected answers to correct the mistakes made by our automatic validation, removing the worker from our blacklist and adding their answers to our gold standard set.

\textbf{Post processing.} We lemmatize all verbs with the WordNet lemmatizer and find the main verb in each state using the VerbNet~\cite{kipper-dissertion-2005}. We detect proper nouns in the collected answers using the DBpedia Spotlight~\cite{daiber-semantics-2013} and replace them with the corresponding common noun, e.g., person names, body parts, animal names, etc. We also remove any possessive determiners for the phrases used in answers.

\begin{table}[t]
\small
\centering
\setlength\tabcolsep{3pt}
\begin{tabular}{|l|c|c|c|c|c|}
\hline
\multicolumn{1}{|c|}{Category}                  & Motion    & Contact   & Percp.    & Body      & Comm.     \\\hline
\multicolumn{1}{|c|}{\multirow{5}{*}{Examples}} & jump      & stand     & look      & smile     & nod       \\
                                                & turn      & touch     & stare     & blink     & point      \\
                                                & shake     & put       & show      & blow      & talk       \\
                                                & run       & open      & hide      & laugh     & wave       \\
                                                & move      & sit       & watch     & wink      & face       \\\hline
LSMDC-QA~\cite{rohrbach-arxiv-2016} & 27.98\%   & 19.09\%   & \textbf{14.78}\%  & 4.43\%   & 5.19\%    \\
MovieQA~\cite{tapaswi-cvpr-2016}    & 13.90\%   & 11.76\%   & 4.95\%    & 2.18\%    & \textbf{12.17}\%  \\
TGIF-QA\texttt{(ours)}              & \textbf{38.04}\% & \textbf{24.78}\% & 9.45\% & \textbf{7.13}\% & 6.78 \% \\\hline
\end{tabular}
\vspace{3pt}
\caption{Distributions of verbs in the answers from different datasets. We show top five most common categories with example verbs. Percp.: perception, comm.: communication.}
\label{tab:verb_distribution_between_datasets} 
\vspace{-6pt}
\end{table}

\begin{table*}[t]
\small
\centering
\setlength\tabcolsep{3pt}
\begin{tabular}{lllllrr}
\hline
Dataset                                 & Objective                                     & Q. Type & Video Source    & Text Source               & \# QA pairs   & \# Clips \\\hline %
LSMDC-QA~\cite{rohrbach-arxiv-2016}     & Fill-in-the-blank for caption completion  & OE        & Movie         & Movie caption             & 348,998       & 111,744   \\      %
MovieQA~\cite{tapaswi-cvpr-2016}        & Visual-textual story comprehension        & MC        & Movie         & Movie synopsis            & 14,944        & 6,771     \\      %
TGIF-QA\texttt{(ours)}                  & Spatio-temporal reasoning from video      & OE \& MC  & Social media  & Caption \& crowdsourced   & 165,165       & 71,741    \\\hline        %
\end{tabular}
\vspace{1pt}
\caption{Comparison of three video VQA datasets (Q.: question, OE: open-ended, and MC: multiple choice).}  %
\label{tab:comparision_between_video_dataset} 
\vspace{-6pt}
\end{table*}

\textbf{QA generation.} We generate QA pairs using the templates shown in Table~\ref{tab:question_template}. It is possible that the generated questions have grammatical errors; we fix those using the LanguageTool. We then generate multiple choice options for each QA pair, selecting four phrases from our dataset. 

Specifically, we represent all verbs in our dictionary as a $300$D vector using the GloVe word embedding~\cite{pennington-emnlp-2014}, pre-trained on the Common Crawl dataset. We then select four verbs, one by one in a greedy manner, whose cosine similarity with the verb from the answer phrase is smaller than the 50th percentile, while at the same time the average cosine similarity from the current set of candidate verbs is minimal -- this encourages diversity in negative answers. We then choose four phrases by maximizing cosine similarity of skip-thought vectors~\cite{kiros-nips-2015} pretrained on the BookCorpus dataset~\cite{zhu-iccv-2015}.

For the repetition counting task, we automatically added samples that had zero count of an action, by randomly pairing a question from our question list with a GIF that was identified as having no repeating action.

\begin{figure*}[tp]
    \centering
    \includegraphics[width=1.0\linewidth]{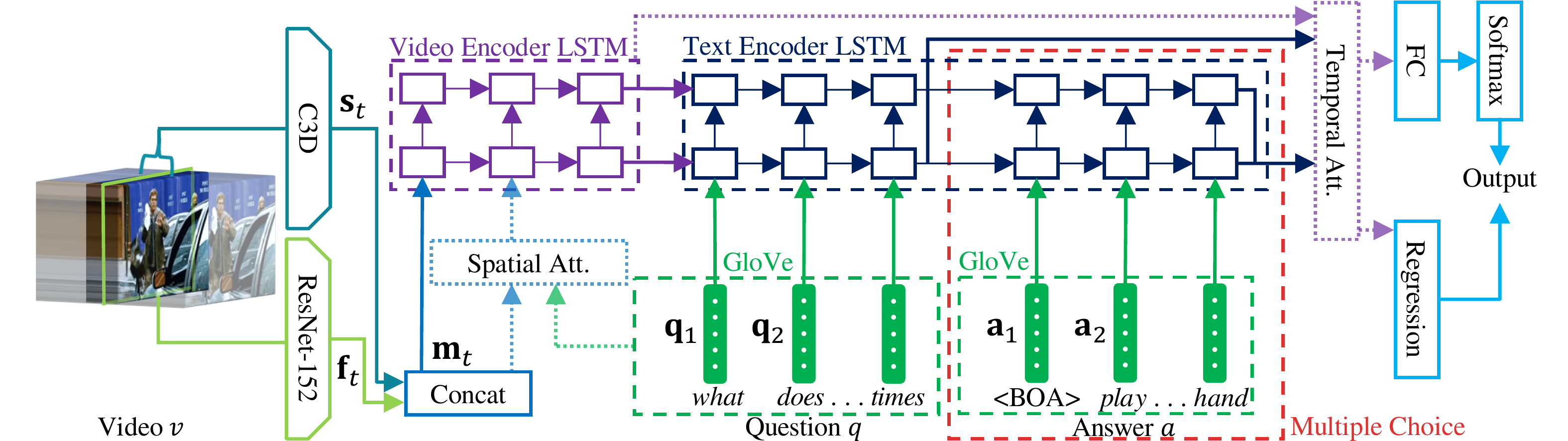}
    \vspace{3pt}
    \caption{The proposed ST-VQA model for spatio-temporal VQA. See Figure~\ref{fig:attention_diagram} for the structure of spatial and temporal attention modules.}
    \label{fig:base_model_structure}
    \vspace{-6pt}
\end{figure*}

\subsection{Comparison with Other Video VQA Datasets} 
\label{sec:dataset_comparision}

Table~\ref{tab:comparision_between_video_dataset} compares our dataset with two existing video VQA datasets. LSMDC-QA refers to the subset of the data used for the VQA task in the LSMDC 2016 Challenge.

It shows that TGIF-QA is unique in terms of the objective and the sources of video and text. \ie, it includes short video clips (GIFs) collected over social media, whereas the other two includes movie clips. Ours also includes both types of questions, open-ended and multiple choice, unlike other datasets. While our dataset is smaller than LSMDC-QA, we include tasks unique to video VQA. Therefore, our dataset can complement existing datasets with unique tasks.

Table~\ref{tab:verb_distribution_between_datasets} shows the distribution of verbs used in answers. We show top five most common verb categories obtained from the WordNet hierarchy. Most notably, TGIF-QA contains more dynamic verbs, such as the ones from the \textit{motion} and the \textit{contact} categories. This is an important characteristic of our dataset because it suggests the need for spatio-temporal reasoning to understand the content. %

\section{Approach}
\label{sec:approach}

We present spatio-temporal VQA (ST-VQA) model for our task (see Figure~\ref{fig:base_model_structure}). The input to our model is a tuple $(v, q, a)$  of a video $v$, a question sentence $q$, and an answer phrase $a$; the answer phrase $a$ is optional and provided only from multiple choice questions (indicated as red dashed box in Figure~\ref{fig:base_model_structure}). The output is either a single word (for open-ended questions) or a vector of compatibility scores (for multiple choice questions). Our ST-VQA model captures visual-textual association between a video and QA sentences using two dual-layer LSTMs, one for each input. 

\subsection{Feature Representation}
\label{sec:input_representation}
\textbf{Video representation}. We represent a video both at the frame-level and at the sequence-level. For the frame features, we use the ResNet-152~\cite{he-cvpr-2016} pretrained on the ImageNet 2012 classification dataset~\cite{russakovsky-ijcv-2015}. For the sequence features, we use the C3D~\cite{tran-iccv-2015} pretrained on the Sport1M dataset~\cite{karpathy-cvpr-2014}. We sample one every four frames to reduce the frame redundancy. For the C3D features, we take 16 subsequent frames centered at each time step, and pad the first or the last frame if too short. We denote the two video descriptors, ResNet-152 and C3D, by $\{\mathbf f_t \}_{t=1}^T$ and $\{\mathbf s_t \}_{t=1}^T$, respectively; $T$ is the sequence length.

Depending on whether we use our spatio-temporal attention mechanism (explained in Section~\ref{sec:attention_mechanism}), we use different feature representations. For the ResNet-152 feature, we take the feature map of the \texttt{res5c} layer ($\in \mathbb R^{7 \times 7 \times 2,048}$) for the spatial attention mechanism and the \texttt{pool5} features ($\in \mathbb R^{2,048}$) for the others. Similarly, for the C3D features, we take the \texttt{conv5b} layer ($\in \mathbb R^{7 \times 7 \times 1,024}$) for the spatial attention mechanism and the \texttt{fc6} feature for the others.

\textbf{Text representation}. There are two types of text inputs: question and answer. A question is a complete sentence, while an answer is a phrase. We simply consider both as a sequence of words and represent them in the same way. For a given input, we represent each word as a $300$D vector using the GloVe word embedding~\cite{pennington-emnlp-2014} pretrained on the Common Crawl dataset. We denote the text descriptor for questions and answers by $\{\mathbf q_n \}_{n=1}^N$ and $\{\mathbf a_m \}_{m=1}^M$, respectively; $N$ and $M$ are the sequence lengths.

\subsection{Video and Text Encoders}
\label{sec:input_encoders}
\textbf{Video encoder.} We encode video features $\{\mathbf s_t \}_{t=1}^T$ and $\{\mathbf f_t \}_{t=1}^T$ using the video encoding LSTM, shown in the purple dashed box in Figure~\ref{fig:base_model_structure}. We first concatenate the features $\mathbf m_t = [\mathbf s_t; \: \mathbf f_t]$, and feed them into the dual-layer LSTM one at a time, producing a hidden state $\mathbf h_{t}^{v} \in \mathbb R^{D}$ at each step:
\begin{align}
    \label{eq:vid_lstm}
    \mathbf h_t^v =  \mbox{LSTM} (\mathbf m_t , \mathbf h_{t-1}^v).
\end{align}
Since we employ a dual-layer LSTM, we obtain pairs of hidden states $\mathbf h_{t}^{v} = ( \mathbf h_{t}^{v,1}, \mathbf h_{t}^{v,2} )$. For brevity, we use the combined form $\mathbf h_{t}^{v}$ for the rest of the paper. We set the dimension $D=512$.

\textbf{Text encoder.} We encode text features of question $\{\mathbf q_n \}_{n=1}^N$ and answer choices $\{\mathbf a_m \}_{m=1}^M$ using the text encoding LSTM, shown in the navy dashed box in Figure~\ref{fig:base_model_structure}. 
While open-ended questions involve only a question, multiple choice questions come with a question and a set of answer candidates. 
We encode a question $\{\mathbf q_n \}_{n=1}^N$ and each of the answer choices $\{\mathbf a_m \}_{m=1}^M$ using a dual-layer LSTM:
\begin{align}
    \label{eq:question_lstm}
    \mathbf h_n^q &=   \mbox{LSTM} (\mathbf q_n, \mathbf h_{n-1}^q ), \:\:\:\: \mathbf h_0^q = \mathbf h_T^v. \\
    \label{eq:answer_lstm}
    \mathbf h_m^a &=   \mbox{LSTM} (\mathbf a_m, \mathbf h_{m-1}^a ), \:\:\:\: \mathbf h_0^a = \mathbf h_N^q
\end{align}
We set the initial hidden state $\mathbf h_0^q$ to the last hidden state of the video encoder $\mathbf h_T^v$, so that visual information is ``carried over'' to the text encoder -- an approach similar to other sequence-to-sequence models~\cite{sutskever-nips-2014, venugopalan-iccv-2015}.
To indicate the starting point of the answer candidate, we put a special character, \texttt{<BOA>} (begin of answer). 
We also use the last hidden state of the question encoder as the initial hidden state of the answer encoder. 
Similar to the video encoder, we set the dimension of all the hidden states to $D=512$.

\subsection{Answer Decoders}
\label{sec:answer_decoder}
We design three decoders that provide answers: one for the multiple choice, the other two for the open-ended.

\textbf{Multiple choice.} We define a linear regression function that takes as input the final hidden states from the answer encoder, $\mathbf h_M^{a}$, and outputs a real-valued score for each answer candidate,
\begin{align}
\label{eq:decoder_mc}
  s =  \mathbf  W_s^{\top} \mathbf h_M^{a} %
\end{align}
where $\mathbf W_s \in \mathbb R^{1,024}$ is the model parameter. We train the decoder by minimizing the hinge loss of pairwise comparisons, $\max(0, 1 + s_{n} - s_{p})$, where $s_n$ and $s_p$ are scores computed from an incorrect and correct answers, respectively. We use this decoder to solve repeating action and state transition tasks.

\textbf{Open-ended, number.} Similar to the above, we define a linear regression function that takes as input the final hidden states from the answer encoder, and outputs an integer-valued answer by adding a bias term $b_s$ to Equation~\eqref{eq:decoder_mc}. We train the decoder by minimizing the $\ell_2$ loss between the answer and the predicted value. We use this encoder to solve the repetition count task.

\textbf{Open-ended, word.} We define a linear classifier that takes as input the final hidden states from the question encoder, $\mathbf h_N^{q} \in \mathbb R^{1,024}$, and selects an answer from a vocabulary of words $\mathcal{V}$ by computing a confidence vector $\mathbf o \in \mathbb R^{|\mathcal V|}$
\begin{align}
\label{eq:decoder_oe}
  \mathbf o = \mbox{softmax} \left ( \mathbf  W_o^{\top} \mathbf h_N^{q} + \mathbf b_o \right )
\end{align}
where $ \mathbf W_o \in \mathbb R^{|\mathcal V|\times 1,024}$ and $\mathbf b_o \in \mathbb R^{|\mathcal V|}$ are model parameters. We train the decoder by minimizing the softmax loss function. The solution is obtained by $y = \argmax_{\mathbf y \in \mathcal V} (\mathbf o)$. We use this encoder to solve the frame QA task.

\subsection{Attention Mechanism} 
\label{sec:attention_mechanism}

\begin{figure}[tp]
    \centering
    \includegraphics[width=\linewidth]{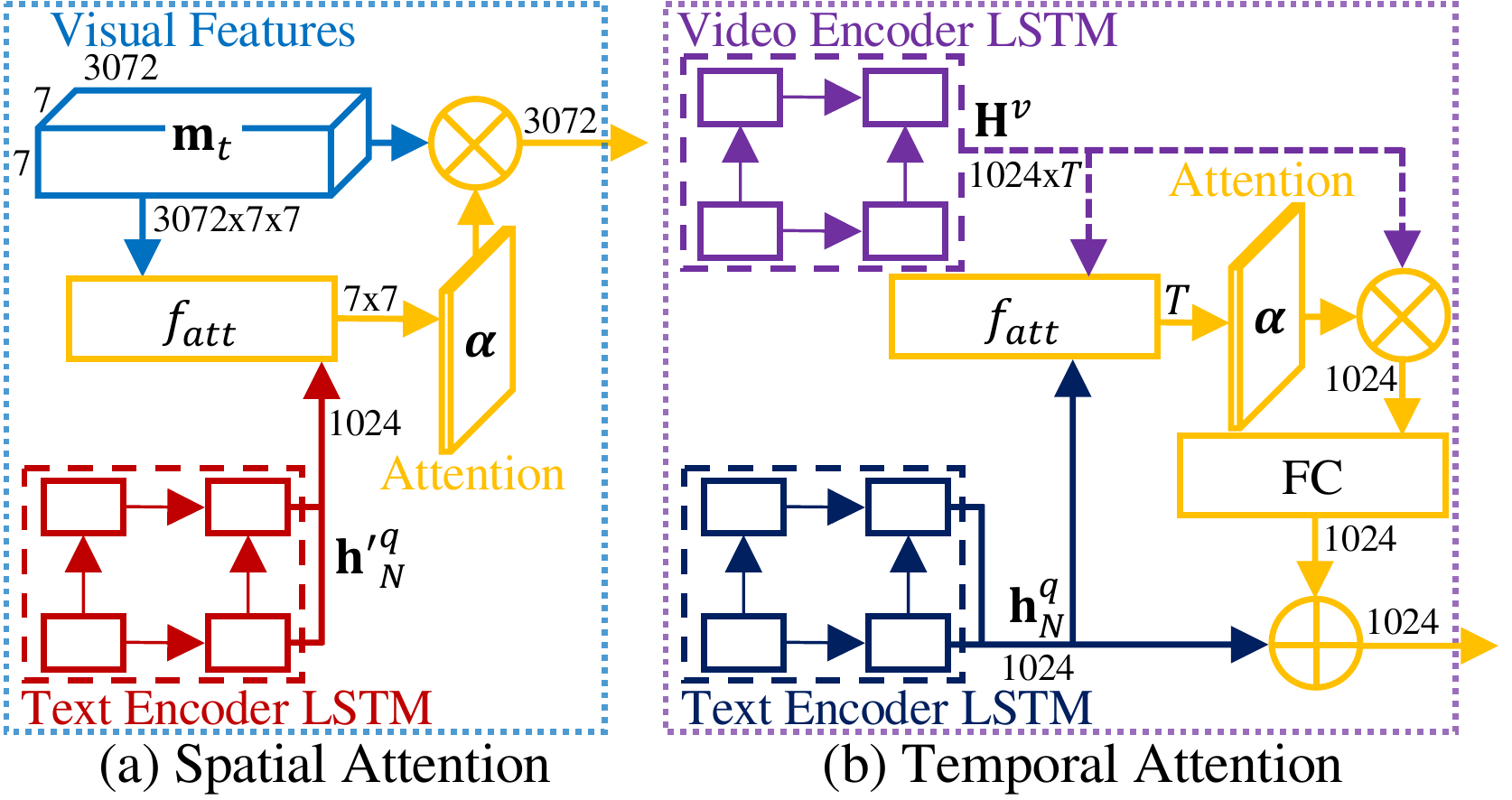}
    \caption{Our spatial and temporal attention mechanisms.}
    \label{fig:attention_diagram}
    \vspace{-6pt}
\end{figure}

While our tasks require spatio-temporal reasoning from videos, the model explained so far is inadequate for such tasks because, in theory, the video encoder ``squashes'' necessary details of the spatio-temporal visual information into a flat representation. We now explain our spatial and temporal attention mechanisms, illustrated in Figure \ref{fig:attention_diagram}. The former allows us to learn \textit{which regions in each frame of a video} to attend to, while the latter allows us to learn \textit{which frames in a video} to attend to solve our tasks. As such, we employ different mechanisms to model each attention type, based on Xu~\etal~\cite{xu-icml-2015} for spatial attention and Bahdanau~\etal~\cite{bahdanau-iclr-2015} for temporal attention. %

\textbf{Spatial attention}. To learn \textit{which regions in a frame} to attend for each word, we use visual representation that preserves spatial information and associate it with a QA pair. Also, we need textual signals when encoding each frame in the video decoder. However, as the model takes a QA pair only after encoding a video, this information is not available a priori. We solve this issue by simply defining another dual-layer LSTM that shares its model parameters with the text encoder. 

Figure~\ref{fig:attention_diagram} (a) illustrates our spatial attention mechanism. For each time step $t$ in a video sequence, we compute a $7 \times 7$ spatial attention mask $\boldsymbol\alpha_{t} = f_{att}(\mathbf{h^\prime}_{N}^{q},\mathbf{m}_{t})$, where $\mathbf{h^\prime}_{N}^{q} \in \mathbb{R}^{1,024}$ is the output of the text encoder and $\mathbf{m}_{t} \in \mathbb{R}^{7 \times 7 \times 3,072}$ is the visual feature map. We then pass the attended visual feature $\boldsymbol\alpha_{t} \mathbf{m}_{t} \in \mathbb{R}^{3,072}$ to the video encoder. The function $f_{att}(\cdot,\cdot)$ is a multi-layer perceptron (MLP) that operates over each of $7 \times 7$ spatial locations, followed by the \texttt{softmax} function. Our MLP is a single layer of 512 hidden nodes with the \texttt{tanh} activation function.

\textbf{Temporal attention}. To learn \textit{which frames in a video} to attend to, we use a visual representation that preserves temporal information and associate it with a QA pair. 

Figure~\ref{fig:attention_diagram} (b) shows our temporal attention mechanism. After we encode video and question sequences, we compute a $1 \times T$ temporal attention mask $\boldsymbol\alpha = f_{att}(\mathbf{h}^{q}_{N}, \mathbf{H}^{v})$, where $\mathbf{h}^{q}_{N} \in \mathbb{R}^{1,024}$ is the last state of the text encoder and $\mathbf{H}^{v} \in \mathbb{R}^{T \times 1,024}$ is a state sequence from the video encoder. We then compute the attended textual signal $\texttt{tanh}(\boldsymbol\alpha \mathbf{H}^{v}\mathbf{W}_{\alpha}) \oplus \mathbf{h}^{q}_{N}$, where $\mathbf{W}_{\alpha} \in \mathbb{R}^{1,024 \times 1,024}$ and $\oplus$ is an element-wise sum, and pass it to the answer decoder. We use the same $f_{att}(\cdot,\cdot)$ as with our spatial attention, with its MLP operating over the temporal dimension $T$.

\subsection{Implementation Details}
\label{sec:impl_detail}

We use the original implementations of ResNet~\cite{he-cvpr-2016}, C3D~\cite{tran-iccv-2015}, and GloVe~\cite{pennington-emnlp-2014} to obtain features from videos and QA text. All the other parts of our model are implemented using the TensorFlow library. Except for extracting the input features, we train our model end-to-end. For the dual-layer LSTMs, we apply layer normalization~\cite{ba-arxiv-2016} to all cells, with the dropout~\cite{pham-icfhr-2014} with a rate of 0.2. 
For training, we use the ADAM optimizer~\cite{kingma-iclr-2015}  with an initial learning rate of 0.001. 
All weights in LSTMs are initialized from a uniform distribution, and all the other weights are initialized from a normal distribution. %

\section{Experiments}
\label{sec:experiments}

We tackle open-ended word and multiple choice tasks as multi-class classification, and use the accuracy as our evaluation metric, reporting the percentage of correctly answered questions. For the open-ended number task, we use the mean $\ell_2$ loss as our evaluation metric to account for the ordinal nature of the numerical labels. We split the data into training and test sets as shown in Table~\ref{tab:dataset}, following the setting in the original TGIF dataset~\cite{li-cvpr-2016}.

\subsection{Baselines}
\label{sec:baseline}

We compare our approach against two recent image-based VQA methods~\cite{fukui-emnlp-2016,ren-nips-2015}, as well as one video-based method~\cite{yu-arxiv-2016}. For fair comparisons, we re-implemented the baselines in TensorFlow and trained them from scratch using the same set of input features. %

\textbf{Image-based}. We select two state-of-the-art methods in image-based VQA: VIS+LSTM~\cite{ren-nips-2015} and VQA-MCB~\cite{fukui-emnlp-2016}. VIS+LSTM combines image representation with textual features encoded by an LSTM, after which it solves open-ended questions using a softmax layer~\cite{ren-nips-2015}. VQA-MCB, on the other hand, uses multimodal compact bilinear pooling to handle visual-textual fusion and spatial attention~\cite{fukui-emnlp-2016}. This model is the winner of the VQA 2016 challenge. 

Since both methods take a single image as input, we adjust them to be applicable to video VQA. We evaluate two simple approaches: \textit{aggr} and \textit{avg}. The \textit{aggr} method aggregates input features of all frames in a video by averaging them, and uses it as input to the model. The \textit{avg} method, on the other hand, solves the question using each frame of a video, one at a time, and report the average accuracy across all frames of all videos, i.e., $1/N \sum_{i=1}^{N} ( 1/M_i \sum_{j=1}^{M_i} \mathbb{I}[y_{i,j} = y_{i}^*] )$, where $N$ is the number of videos, $M_i$ is the number of frames for the $i$-th video, $\mathbb{I}[\cdot]$ is an indicator function, $y_{i,j}$ is a predicted answer for the $j$-th frame of the $i$-th video, and $y_{i}^*$ is an answer for the $i$-th video.

\begin{table}[t]
\renewcommand{\arraystretch}{1.1}
\small
\centering
\begin{tabular}{|c|c|c|c|c|c|}
\hline
\multicolumn{2}{|c|}{\multirow{2}{*}{Model}}    & \multicolumn{2}{|c|}{Repetition} & State & Frame   \\\cline{3-4}
\multicolumn{2}{|c|}{}              & Count  & Action & Trans. & QA                       \\\cline{1-6} %
\multicolumn{2}{|l|}{Random chance} & 6.9229 & 20.00 & 20.00 & 0.06 \\\cline{1-6}      %
VIS+LSTM                    & aggr  & 5.0921 & 46.84 & 56.85 & 34.59 \\
  \cite{ren-nips-2015}      & avg   & 4.8095 & 48.77 & 34.82 & 34.97 \\\cline{1-6}
VQA-MCB                     & aggr  & 5.1738 & 58.85 & 24.27 & 25.70 \\
\cite{fukui-emnlp-2016}     & avg   & 5.5428 & 29.13 & 32.96 & 15.49 \\\cline{1-6}
\multicolumn{2}{|l|}{Yu~\etal~\cite{yu-arxiv-2016}} & 5.1387 & 56.14 & 63.95 & 39.64 \\\cline{1-6}
\multicolumn{2}{|l|}{ST-VQA-Text}  & 5.0056 & 47.91 & 56.93 & 39.26 \\
\multicolumn{2}{|l|}{ST-VQA-ResNet} & 4.5539 & 59.04 & 65.56 & 45.60 \\
\multicolumn{2}{|l|}{ST-VQA-C3D}    & 4.4478 & 59.26 & 64.90 & 45.18 \\
\multicolumn{2}{|l|}{ST-VQA-Concat} & \underline{4.3759} & \underline{60.13} & \underline{65.70} & \underline{48.20} \\\cline{1-6}
\multicolumn{2}{|l|}{ST-VQA-Sp.}    & \textbf{4.2825} & 57.33 & 63.72 & 45.45 \\
\multicolumn{2}{|l|}{ST-VQA-Tp.}    & 4.3981 & \textbf{60.77} & \textbf{67.06} & \textbf{49.27} \\
\multicolumn{2}{|l|}{ST-VQA-Sp.Tp.} & 4.5614 & 56.99 & 59.59 & 47.79 \\
\hline
\end{tabular}
\vspace{10pt}
\caption{Experimental results of VQA according to different problem types on our TGIF-QA dataset. (Sp.) indicates the spatial attention and (Tp.) means temporal one.
We report the mean $\ell_2$ loss for the repetition count task, and the accuracy for the other three tasks.} 
\label{tab:experiment_results} 
\vspace{-5pt}
\end{table}

\begin{figure*}[t!]
\centering
\vspace{-5pt}
\includegraphics[width=0.98\linewidth]{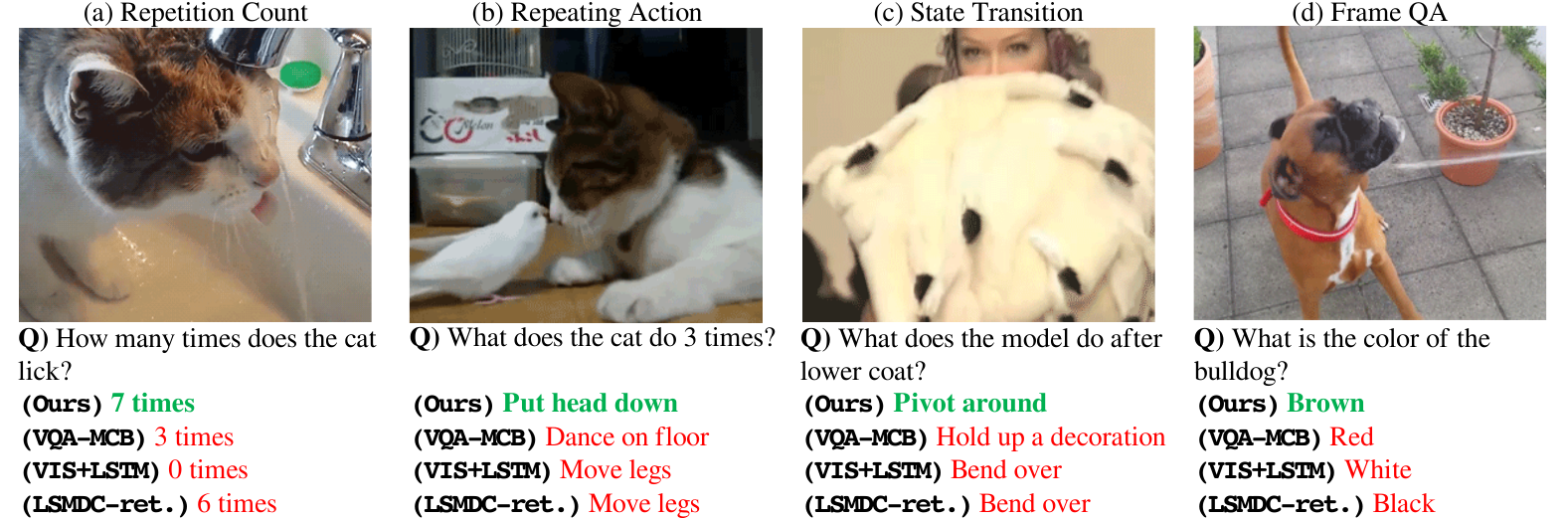}
\caption{Qualitative comparison of VQA results from different approaches, on the four task types of our TGIF-QA dataset.}
\label{fig:examp_vidqa}
\vspace{-6pt}
\end{figure*}

\textbf{Video-based}. We select the state-of-the-art method in video VQA, Yu~\etal~\cite{yu-arxiv-2016}, which has won the retrieval track in the LSMDC 2016 benchmark. We use their retrieval model that employs the same decoder as explained in section \ref{sec:answer_decoder}. Although the original method used an ensemble approach, we here use a single model for a fair comparison.

\textbf{Variants of our method}. To conduct an ablation study of our method, we compare seven variants of our model, as shown in Table~\ref{tab:experiment_results}. The four (Text, ResNet, C3D, Concat) compare different representations for the video input; Text uses neither ResNet nor C3D features, whereas Concat uses both ResNet and C3D features. They also do not employ our spatial and temporal attention mechanisms. The next two variants (Spatial and Temporal) include either one of the attention mechanisms. Finally, we evaluate a combination of the two attention mechanisms, by training the temporal part first and finetuning the spatial part later. %

\subsection{Results and Analysis}
\label{sec:results_and_analysis}

Table~\ref{tab:experiment_results} summarizes our results. We observe that video-based methods outperform image-based methods, suggesting the need for spatio-temporal reasoning in solving our video QA tasks. We note, however, that the differences may not be seen significant; we believe this is because the C3D features already capture spatio-temporal information to some extent. 

A comparison between different input features of our method (Text, ResNet, C3D, Concat) suggests the importance of having both visual representations in our model. Among the four baselines, the Concat approach that uses both features achieves the best performance across all tasks. 

A comparison between different attention mechanisms (Spatial and Temporal) shows the effectiveness of our temporal attention mechanism, achieving the best performance in three tasks. Similar results are reported in the literature; for example, in the video captioning task, Yao~\etal~\cite{yao-iccv-2015} obtained the best result by considering both local and global temporal structures.

Finally, Figure \ref{fig:examp_vidqa} shows some qualitative examples from different approaches on the four task types of TGIF-QA. We observe that answering the questions indeed requires spatio-temporal reasoning. For example, the cat in Figure~\ref{fig:examp_vidqa} (b) puts head down multiple times, which cannot be answered without spatio-temporal reasoning. Our method successfully combines spatial and temporal visual representation from the input-level via ResNet and C3D features, and learns to selectively attend to them via our two attention mechanisms.

\section{Conclusion}
\label{sec:conclusion}

Our work complements and extends existing work on VQA with three main contributions: (i) proposing three new tasks that require spatio-temporal reasoning from videos, (ii) introducing a new large-scale dataset of video VQA with 165K QA pairs from 72K animated GIFs, and (iii) designing a dual-LSTM based approach with both spatial and temporal attention mechanisms.

Moving forward, we plan to improve our ST-VQA model in several directions. Although our model is based on a sequence-to-sequence model~\cite{venugopalan-iccv-2015} to achieve simplicity, it can be improved in different ways, such as adopting the concept of 3D convolution~\cite{tran-iccv-2015}.
Another direction is to find better ways to combine visual-textual information. Our model without the attention module (\eg Concat in Table~\ref{tab:experiment_results}) combines visual-textual information only at the text encoding step. Although our attention mechanisms explored ways to combine the two modalities to some extent, we believe there can be more principled approaches to do it efficiently, such as the recently proposed multimodal compact bilinear pooling~\cite{fukui-emnlp-2016}. %

\section{Document Changelog}
\label{sec:document changelog}
To help readers understand how it had changed over time, here's a brief changelog describing the revisions.
\begin{compactitem}
\item[]\textbf{v1}~(Initial) CVPR 2017 camera-ready version.
\item[]\textbf{v2}~Added statistics and results, including text-only baseline, for extended dataset.
\item[]\textbf{v3}~Updated the results in Table~\ref{tab:experiment_results} and uploaded relevant files to our repository.
\end{compactitem}

\medskip\textbf{Acknowledgements.}
We thank Miran Oh for the discussions related to natural language processing, as well as Jongwook Choi for helpful comments about the model.
We also appreciate Cloud \& Mobile Systems lab and Movement Research lab at Seoul National University for renting a few GPU servers for this research. 
This work is partially supported by Big Data Institute (BDI) in Seoul National University and Academic Research Program in Yahoo Research.
Gunhee Kim is the corresponding author. 

{\small
\bibliographystyle{ieee}
\bibliography{cvpr17_tgifqa}
}

\end{document}

%% file: mymacros.tex
\usepackage{color}
\usepackage{amsmath}
\usepackage{amssymb}
\usepackage{multirow, varwidth}
\usepackage{epsfig}
\usepackage{verbatim}
\makeatletter
\newif\if@restonecol
\makeatother

\usepackage[ruled,vlined]{algorithm2e}
\usepackage{xr}
\usepackage[amsmath,thmmarks]{ntheorem}
\usepackage{flushend}

\theorembodyfont{\normalfont}
\theoremseparator{.}

\theoremstyle{nonumberplain}

\newcommand{\argmax}{\operatornamewithlimits{argmax}}

\DeclareRobustCommand\onedot{\futurelet\@let@token\@onedot}
\def\onedot{.} 
\def\eg{\emph{e.g}\onedot} 
\def\ie{\emph{i.e}\onedot}

\def\etal{\emph{et al}\onedot}